\documentclass{article}

\usepackage{arxiv}

\usepackage[utf8]{inputenc} %
\usepackage[T1]{fontenc}    %
\usepackage[hyphens]{url}   %
\usepackage{hyperref}       %
\usepackage{booktabs}       %
\usepackage{tabularx}
\usepackage{amsfonts}       %
\usepackage{wrapfig}
\usepackage{microtype}      %
\usepackage{graphicx}
\usepackage{natbib}
\usepackage{doi}
\usepackage[table,dvipsnames]{xcolor}
\usepackage{authblk}
\usepackage[table-format=2.2, mode=match, detect-all=true]{siunitx}
\usepackage[section]{placeins} %
\usepackage{multirow}
\usepackage{soul} %
\usepackage{xspace}  

\hypersetup{
    colorlinks,
    linkcolor={blue!50!black},
    citecolor={blue!50!black},
    urlcolor={blue!50!black}
}

\usepackage[nolist]{acronym}

\newcommand\blfootnote[1]{%
  \begingroup
  \renewcommand\thefootnote{}\footnote{#1}%
  \addtocounter{footnote}{-1}%
  \endgroup
}

\usepackage{cleveref}
\crefformat{section}{\S#2#1#3}
\crefformat{subsection}{\S#2#1#3}
\crefformat{subsubsection}{\S#2#1#3}
\crefrangeformat{section}{\S\S#3#1#4 to~#5#2#6}
\crefmultiformat{section}{\S\S#2#1#3}{ and~#2#1#3}{, #2#1#3}{ and~#2#1#3}
\usepackage{refstyle}
\Crefformat{figure}{#2Fig.~#1#3}
\Crefmultiformat{figure}{Figs.~#2#1#3}{ and~#2#1#3}{, #2#1#3}{ and~#2#1#3}
\Crefformat{table}{#2Table~#1#3}
\Crefmultiformat{table}{Tabs.~#2#1#3}{ and~#2#1#3}{, #2#1#3}{ and~#2#1#3}
\Crefformat{appendix}{Appx.~\S#2#1#3}
\crefformat{algorithm}{Alg.~#2#1#3}

\title{MAIRA-1: A specialised large multimodal model for radiology report generation}

\date{} 					%

\author[1]{Stephanie L. Hyland*}
\author[1]{Shruthi Bannur}
\author[1]{Kenza Bouzid}
\author[1]{Daniel C. Castro}
\author[2]{Mercy Ranjit}
\author[1]{Anton Schwaighofer}
\author[1]{Fernando Pérez-García}
\author[1]{Valentina Salvatelli}
\author[2]{Shaury Srivastav}
\author[1]{Anja Thieme}
\author[3]{Noel Codella}
\author[4]{Matthew P. Lungren}
\author[1]{Maria Teodora Wetscherek}
\author[1]{Ozan Oktay}
\author[1]{Javier Alvarez-Valle*}
\affil[1]{Health Futures, Microsoft Research}
\affil[2]{Microsoft Research India}
\affil[3]{Microsoft Azure AI}
\affil[4]{Microsoft Health and Life Sciences}

\renewcommand{\shorttitle}{\maira{}}

\newcommand\WIP[1]{\bgroup\color{blue}#1\egroup}

\begin{document}
    \begin{acronym}
    \acro{VQA}{visual question answering}
    \acro{LLM}{large language model}
    \acro{SOTA}{state of the art}
    \acro{NPV}{negative predictive value}
    \acro{PPV}{positive predictive value}
    \acro{RL}{reinforcement learning}
    \acro{PA}{posteroanterior}
    \acro{AP}{anteroposterior}
    \acro{CXR}{chest X-ray}
\end{acronym}
    \maketitle
    \newcommand{\domainimageencoder}{RAD-DINO\xspace}
\newcommand{\maira}{MAIRA-1}
\newcommand{\modelname}[1]{\texttt{#1}}

\newcommand{\reportsection}[1]{\textit{#1}}
\newcommand{\findings}{\reportsection{Findings}}
\newcommand{\impression}{\reportsection{Impression}}
\newcommand{\indication}{\reportsection{Indication}}

\newcommand{\ctext}[2]{{\sethlcolor{#1}\hl{#2}}}
\newcommand{\reporterror}[1]{\ctext{red!20}{#1}}
\newcommand{\reportomission}[1]{\ctext{lightgray!50}{[\textit{#1}]}}
\newcommand{\reporthighlight}[1]{\ctext{ProcessBlue!20}{#1}}

    \begin{abstract}
We present a radiology-specific multimodal model for the task for generating radiological reports from \acp{CXR}. Our work builds
on the idea that \acp{LLM} can be equipped with multimodal capabilities through alignment with pre-trained vision encoders. On natural images,
this has been shown to allow multimodal models to gain image understanding and description capabilities.
Our proposed model (\maira{}) leverages a CXR-specific image encoder in conjunction with a fine-tuned \ac{LLM} based on Vicuna-7B, and text-based data augmentation, to produce reports with state-of-the-art quality. In particular, \maira{}  significantly improves on the radiologist-aligned RadCliQ metric and across all lexical metrics considered.
Manual review of model outputs demonstrates 
promising fluency and accuracy of generated reports while uncovering failure modes not captured by existing evaluation practices.
More information and resources can be found on the project website: \url{https://aka.ms/maira}. \blfootnote{*Corresponding authors: \url{sthyland@microsoft.com}, \url{jaalvare@microsoft.com}}

\end{abstract}

    \section{Introduction}
Large-scale pretraining of general-purpose image and language models has enabled the development of data-efficient large multimodal models. A common paradigm is to adapt a vision encoder to a pretrained \ac{LLM} with varying levels of integration, as done by LLaVA~\citep{liu_visual_2023}, InstructBLIP~\citep{Dai2023InstructBLIPTG} and Flamingo~\citep{flamingo}. Here, we explore this paradigm for the specialised goal of generating radiology reports from images, namely generating the \findings{} section of a report based on a frontal chest X-ray and given the \indication{} for the study.

The \findings{} section contains the radiologist's key observations about the image. To assist in their interpretation, a radiologist may refer to prior scans from the patient, other imaging modalities or views of the chest, or the patient's clinical and medical history if available and not provided in the \indication{}. Hence, reporting on a chest X-ray requires synthesis of various data sources including multiple images taken over a period of time. For this study, we focus on the simplified single-image setting, acknowledging that the resulting model will be susceptible to generating `hallucinated' references to prior studies~\citep{ramesh2022improving,bannur2023learning} or lateral views.

Although the \findings{} section aims describe the image, the generation task crucially differs from image  captioning.
Chest X-ray reports must also establish the \emph{absence} of findings, for example confirming that the insertion of a central venous line did not cause a pneumothorax.  Secondly, the findings in a chest X-ray are not typical `objects'. Radiographic findings can be subtle variations in opacity against an otherwise-normal background of overlapping structures, requiring the extraction of fine-grained details from the image. Hence, findings generation remains a challenging multimodal task, requiring both the extraction of fine-grained details from the image and the generation of nuanced, radiology-specific language. A model which could generate the first draft of a radiology report has the potential to improve and expedite radiology reporting workflows~\citep{huang2023generative}, if it can first demonstrate a sufficient standard of clinical accuracy.

General-domain models demonstrably fail at the task of findings generation~\citep{medpalm}. We therefore develop a \emph{radiology-specific} large multimodal model, which we call \maira{}. \maira{} benefits from a pre-trained language model (Vicuna-7B \citep{vicuna2023}), a radiology-specific image encoder (\domainimageencoder \citep{perez2024rad}), and the use of GPT-3.5 for data augmentation. By fine-tuning this model on a publicly available dataset of chest X-rays with corresponding reports (MIMIC-CXR \citep{johnson2019mimic-cxr,johnson2019mimic-cxr-dataset}), we demonstrate the potential of large multimodal models for specialised radiology use-cases.

Concretely, we share the following outcomes:
\begin{enumerate}
    \item Performance competitive with existing state-of-the-art is possible without extremely large models or datasets.
    \item Design choices make a difference: keys to the success of this model are the use of a domain-specific image encoder with a larger image resolution and a more complex adapter layer, as well as the use of data augmentation and the \emph{Indications} section of the report.
    \item Evaluation for this task remains challenging with heterogeneity across prior work. We report a wide variety of metrics to enable comparison, and demonstrate through stratified analyses the sensitivity of performance measures to test set characteristics.
\end{enumerate}
We share examples of \findings{} sections generated by \maira{} to highlight both its successes and limitations. By re-purposing pre-trained \acp{LLM} we have demonstrated what is possible in a constrained setting with a limited dataset.  
Relative to a radiologist, \maira{} receives only a limited view of the patient, relying on a single chest X-ray and the \indication{} for the study.
Hence, \maira{} is a step \emph{towards} realistic report-drafting systems, which will incorporate richer inputs such as previous images or other clinical information. We anticipate that with larger and cleaner training datasets, this flexible \ac{LLM}-based approach can yield further gains.

    \section{Related work}

\paragraph{Generalist foundation models}

The paradigm of aligning image encoders to existing \acp{LLM} has achieved great success in the general domain. An example is LLaVA~\citep{liu_visual_2023}, which combines a vision encoder with \ac{LLM} using a simple adaptation layer applied to the image embedding, and employing visual instruction tuning to achieve general-purpose visual and language understanding. InstructBLIP~\citep{Dai2023InstructBLIPTG} performs similar visual instruction tuning, but leverages a more powerful Q-former and a frozen \ac{LLM}. Flamingo~\citep{flamingo} couples the image and text more tightly, allowing a frozen \ac{LLM} to cross attend to images. Flamingo additionally benefits from training on interleaved visual-text-video data. PaLM-E \citep{palme} is a multimodal decision making model, trained on text, image and sensor data, capable of embodied reasoning and decision making.   %

\paragraph{Medical/radiology adaptation of LLMs/MLLMs}

There have been multiple efforts to specialise generalist foundation models to the medical domain or specifically for radiology applications.
For example, Med-Flamingo \citep{medflamingo} is based on OpenFlamingo \citep{openflamingo, flamingo} and was trained on images and captions from medical textbooks to perform few-shot visual question answering (VQA).
Med-PaLM M \citep{medpalm} fine-tuned the proprietary PaLM-E model \citep{palme} on a broad collection of biomedical datasets, to build a system addressing diverse uni- and multimodal tasks over text, images, and genomics.
LLaVA-Med \citep{li_llava-med_2023} proposed to adapt LLaVA \citep{liu_visual_2023} with a curriculum of plain image--text pairs and generated multimodal instructions based on PubMed data.
ELIXR \citep{elixr} aligns a CXR encoder model, SupCon \citep{supcon}, with PaLM 2-S \citep{anil2023palm} through a multi-stage training process. The resulting model can perform tasks such as classification, semantic search, question answering and quality assurance.
Radiology-GPT \citep{radiologygpt} is a text-only model based on the Alpaca instruction-tuning framework \citep{alpaca}, trained with radiology reports from the MIMIC-CXR dataset to perform findings-to-impression generation. 

\paragraph{Radiology report generation} Given the long tail of possible observations in a chest X-ray, and the need for fine-grained description of findings, the generation of the narrative radiology report itself from the image is a promising target for machine learning systems~\citep{wang2018tienet}. Such systems necessarily require a generative language modelling component; initially recurrent neural networks~\citep{wang2018tienet,clinically-accurate} giving way more recently to transformer architectures~\citep{miura,r2gen,bannur2023learning}, including \acp{LLM} such as PaLM~\citep{medpalm} and here Vicuna-7B \citep{vicuna2023}.

Given the need for clinical accuracy in the generated text, 
a line of work departs from a vanilla language-modelling loss and uses \ac{RL} to optimise for `clinically relevant' rewards, based on the presence of specific findings~\citep{clinically-accurate,chexpert}, or logical consistency~\citep{miura,semantic_rewards}. A downside of such approaches is the reliance on models such as CheXbert~\citep{chexbert} or RadGraph~\citep{radgraph} to extract clinical entities, and the more complex optimisation problem. Here we demonstrate what is possible through plain auto-regressive language modelling alone, acknowledging that gains from more sophisticated training objectives or \ac{RL}-based approaches are likely complementary.

Prior work has focused on generating different sections of the radiology report, sometimes the combination of both \findings{} and \impression{}~\citep{wang2018tienet,r2gen,promptmrg,style_aware}, the \impression{} alone~\citep{cxr_repair,bannur2023learning}, or---as here---the \findings{} only~\citep{miura,semantic_rewards,rgrg,nicolson,medpalm}. \cite{xrem} and \cite{evaluating_progress} studied all three settings, providing evidence that the choice of section markedly impacts reported metrics, prohibiting comparison between variations of the task.

    \newcommand{\gpt}{\modelname{gpt-3.5-turbo}}

\section{Method}
\label{sec:methods}

In this section, we explain details of our task definition, dataset composition and preparation, modelling choices, and training and inference pipelines. We also describe our broad suite of evaluation metrics.

\subsection{Task and data}

\paragraph{Task description}
We generate the main body section (\findings{}) of the report accompanying a chest X-ray. The \findings{} section contains the radiologist's description of the normal and abnormal findings on the image. The image is typically accompanied by an \indication{} section providing the reason for the study, which may include some clinical history or a specific request from the referring clinician.

The most common chest X-ray view is a frontal image, either acquired from posterior to anterior (PA) or anterior to posterior (AP). Other imaging views of the chest are also routinely performed to aid in the diagnostic task, such as a lateral image. It is worth noting that, in isolation, the lateral view is not able to visualise the relevant anatomy with sufficient clarity to generate a comprehensive diagnostic report. Therefore, the frontal view remains the default view for the traditional chest X-ray clinical interpretation task(s) and our work similarly relies on the frontal view in line with most prior studies.\footnote{A notable exception is \cite{medpalm}, where the lateral view is used.}

In addition, radiology reports typically contain a final \impression{} section summarising the actionable insights from the study, including suspected clinical diagnoses and recommendations for follow-up investigations.
Such information cannot be fully gathered from the image alone---and often not even from the remainder of the report in isolation---relying heavily on the radiologist's domain expertise, external data (e.g.\ patient history), and context of the study.

Hence, we focus on the task of generating specifically the \findings{} section of the report, given the \indication{} section (if available), and a single frontal chest X-ray image.

\paragraph{Dataset description}

\begin{wraptable}[9]{r}{0.34\textwidth}
    \vspace{-\intextsep}
    \vspace{-\abovecaptionskip}
    \centering
    \caption{Number of subjects, studies and DICOMs in each of our splits of the MIMIC-CXR dataset.}
    \footnotesize
    \vspace{\abovecaptionskip}
    \begin{tabular}{@{} lrrr @{}}
        \toprule
        & Train & Validation & Test \\
        \midrule
        Subjects & 55,218 & 2,709 & 285 \\
        Studies & 131,613 & 6,471 & 2,210 \\
        DICOMs & 146,909 & 7,250 & 2,461 \\
        \bottomrule
    \end{tabular}
    \label{tab:mimic_counts}
\end{wraptable}
We train and evaluate on the MIMIC-CXR dataset \citep{johnson2019mimic-cxr,johnson2019mimic-cxr-dataset}, hosted on PhysioNet \citep{goldberger2000physiobank}. This dataset from the Beth Israel Deaconess Medical Center in Boston, comprises a total of 377,110 DICOM images across 227,835 studies. Each imaging study is accompanied by a report. We process the DICOM images to remove all non-AP/PA scans. For each report, we extract the \findings{} and \indication{} sections, using the official MIMIC-CXR codebase.%
    \footnote{\url{https://github.com/MIT-LCP/mimic-cxr/blob/master/txt/section_parser.py}}
We discard all studies for which \findings{} could not be extracted, but we allow for missing \indication{} sections. We use the standard MIMIC-CXR test split in our experiments, however, we additionally exclude benchmark datasets MS-CXR \citep{mscxr} and MS-CXR-T \citep{mscxrt} from our training set. \Cref{tab:mimic_counts} describes the subject, study and DICOM counts in each split.  To facilitate comparison, we provide the list of DICOM identifiers used in our test split as an ancillary file.

We derive our images from the original DICOM files, instead of the commonly used JPEG files from MIMIC-CXR-JPG \citep{johnson2019mimic-cxr-jpg}, the latter of which may contain compression artefacts and further loss of detail from grayscale quantisation.
Following the pre-processing pipeline from CLIP \citep{radford2021clip}, images are resized to match the shortest edge to the input size of the image encoder, then the longest edge is centre-cropped to the same size.
Intensities are normalised according to the image encoder's training data.

\paragraph{Data augmentation using GPT} We use GPT-3.5 to paraphrase both the \findings{} and \indication{} sections of the training set as a data augmentation technique, resulting in an additional 131,558 reports.%
    \footnote{We used a private, compliant deployment of OpenAI's \gpt{} version 0301 on Microsoft Azure (\url{https://learn.microsoft.com/en-us/azure/ai-services/openai/overview}), such that MIMIC data was not sent to public-facing servers.}
We instruct GPT-3.5 to rewrite the findings and indication while preserving the information and radiology style. \Cref{tab:gpt_example} shows an example.
Note that we leave the validation and test sets unchanged.
\begin{table}
    \caption{Example of the output of GPT-3.5 paraphrasing on the \indication{} and \findings{} sections. Paraphrasing preserves the clinical content while introducing small variations in phrasing.}
    \label{tab:gpt_example}
    \begin{tabular}{@{}lp{14.1cm}@{}}\toprule
    Original & INDICATION: \_F with presyncope. r/o infection // ?pneumonia \\
    & FINDINGS: AP and lateral chest radiograph demonstrates hyperinflated lungs. Cardiomediastinal and hilar contours are within normal limits. There is no pleural effusion or pneumothorax. No evidence of pulmonary edema. Lungs are without a focal opacity worrisome for pneumonia. There is no air under the right hemidiaphragm.\\
    \midrule

    Paraphrased & INDICATION: Female patient with pre-syncope. Suspected infection, possibly pneumonia.\\
        & FINDINGS:  Chest radiographs show lungs with hyperinflation, but normal cardiomediastinal and hilar contours. No pleural effusion, pneumothorax, or pulmonary edema is observed. No focal opacity is present in the lungs that could indicate pneumonia. Additionally, there is no air under the right hemidiaphragm.\\
    \bottomrule
    \end{tabular}

\end{table}

\subsection{Model architecture}
Our model consists of an  image encoder, a learnable adapter on top of the image features, and an \ac{LLM}, following the LLaVA-1.5 architecture~\citep{liu_visual_2023,liu_improved_2023}. We use the radiology-specialised image encoder, \domainimageencoder ~\citep{perez2024rad}. \domainimageencoder  is a ViT-B model \citep{dosovitskiy2020vit} with 87M parameters trained on 838k chest X-ray images. The input resolution is 518$\times$518 and the encoder patch size is 14.
We employ Vicuna-7B \citep{vicuna2023} as the LLM. We note from LLaVA-Med \citep{li_llava-med_2023} that this initialisation from an LLM pretrained only on language should lead to better performance than from one already trained on multimodal data.
Our adapter is a multi-layer perceptron (MLP) with GELU activations \citep{hendrycks2016gaussian} and hidden size 1024 for all layers. The adapter weights are randomly initialised following the defaults from PyTorch v2.0.1.

We prompt the model with a system message and a human instruction interleaved with the relevant image. The model is trained to output the correct answer, i.e.\ the full \findings{} section of the report. We first embed the image into a sequence of image patch tokens using the image encoder, taking the embeddings from the last layer of the model and excluding the \texttt{CLS} token. The adapter MLP is applied to each embedding to transform these image tokens to the input space of the LLM. After tokenising and embedding the prompt and answer, we insert the image patch tokens at the specified location in the prompt (typically between the system message and human instruction).
The instructions we use with  \maira{} are ``\{image placeholder\} Provide a description of the findings in the radiology image given the following indication: \{indication\}'' if we know the \indication{} for the image, otherwise ``\{image placeholder\} Provide a description of the findings in the radiology image.''.

\subsection{Training and inference}
We train with a standard auto-regressive language modelling loss (cross-entropy) \citep{graves2013generating}. We use similar hyperparameters to LLaVA-1.5 for training, i.e.\ tuning the LLM jointly with the randomly initialized adapter \citep{liu_visual_2023}. Unlike LLaVA-1.5 we do not have a precursor training step to pretrain the adapter (see \Cref{sec:pretrain_adapter} for an experimental comparison). We train for 3 epochs without any parameter-efficient fine-tuning techniques. We use a cosine learning rate scheduler with a warm-up of 0.03 and learning rate \num{2e-5}. The global batch size is 128. Based on the behaviour of validation metrics observed throughout training in our experiments, we take the final checkpoint for all runs.
For inference, we decode in 32-bit precision up to 150 tokens.

\subsection{Evaluation metrics}
We evaluate the generated reports using both lexical and radiology-specific metrics, as described below. Radiology-specific metrics typically focus on how particular findings are described in the text, for example whether `consolidation' is noted as present or absent. For this reason, they are more sensitive to clinically-relevant aspects of the generated report, rather than potentially-superficial variation in phrasing.

\paragraph{Lexical metrics}

We employ a collection of traditional NLP metrics designed to quantify word overlap between generated and reference texts.
In particular, ROUGE-L \citep{lin_rouge_2004} quantifies the length of the longest common word subsequence relative to the lengths of predicted and reference reports.
BLEU-4 \citep{papineni_bleu_2002} is based on $n$-gram precision (geometric mean for $n$ up to 4), with a brevity penalty to discourage too short predictions.
Lastly, we report METEOR \citep{banerjee_meteor_2005}, which first aligns individual words (unigrams) in the prediction and reference, while attempting to preserve their ordering; then computes a weighted harmonic mean of unigram precision and recall, with a penalty for fragmentation of consecutive word subsequences.%
    \footnote{We run METEOR with the default parameters as originally presented by \citet{banerjee_meteor_2005}.}

\paragraph{CheXpert F1}
This set of metrics uses the CheXbert automatic labeller \citep{chexbert} to extract `present'/`absent'/`uncertain' labels for each of the 14 CheXpert pathological observations \citep{chexpert} from a generated report and the corresponding reference.
As done originally by \citet{chexpert}, we compute two versions of this metric, mapping the `uncertain' label to negative or positive, to enable comparison with papers that use either.
The binary F1 score for each CheXpert category is then computed conventionally as the unweighted harmonic mean of precision and recall.
We report the macro- and micro-averaged F1 scores over 5 major observations%
    \footnote{Following \citet{miura, medpalm}, and others, based on the CheXpert `competition tasks' selected by \cite{chexpert}, the subset of 5 major categories considered is: atelectasis, cardiomegaly, consolidation, edema, and pleural effusion.}
and all 14 observations, referring to them respectively as `[Macro/Micro]-F1-[5/14]'.

\paragraph{CheXbert vector similarity}
This metric feeds the generated and reference reports through the CheXbert model \citep{chexbert}, then calculates the cosine similarity between their embeddings \citep{evaluating_progress}.
We compute this metric as well as RadGraph F1 and RadCliQ using the code released by \citet{evaluating_progress}.%
\footnote{\url{https://github.com/rajpurkarlab/CXR-Report-Metric/tree/v1.1.0}}

\paragraph{RadGraph-based metrics}
The RadGraph model \citep{radgraph} parses radiology reports into graphs containing clinical entities (references to anatomy and observations) and relations between them.
Introduced in \citet{evaluating_progress}, the RadGraph F1 metric computes the overlap in entities and relations separately, then reports their average. Entities are considered to match if the text spans and assigned types are the same, and relations are matched if their endpoint entities and relation type are the same.

We also compute a variant F1 metric used in prior work \citep{semantic_rewards}, to enable direct comparison.
Specifically, we report their RG\textsubscript{ER} score, which matches entities based on their text, type, and whether or not they have at least one relation. For this, we use the \texttt{radgraph} package.\footnote{\url{https://pypi.org/project/radgraph/}}

\paragraph{RadCliQ}
Also proposed by \citet{evaluating_progress_medrxiv}, RadCliQ (Radiology Report Clinical Quality) is a composite metric that integrates RadGraph F1 and BLEU score in a linear regression model to predict the total number of errors that radiologists would identify in a report. In their study, this metric had the closest alignment with radiologists' judgement of report quality.  For consistency with prior work, we use version 0 of RadCliQ.

    \section{Experiments}

In this section, we present an overview of the experimental design, results and ablations. The main experimental design is seen in \Cref{tab:describe_experiments} where we detail design choices related to data, pretraining, image encoders and adapters.
To account for slight differences in the definitions of the test split in other studies (see \Cref{sec:sota_comparison}), for all our experiments we report medians and 95\% confidence intervals, estimated over 500 bootstrap samples of the MIMIC-CXR test set.
For all metrics, higher values are better, with the exception of RadCliQ (indicated by `$\downarrow$' in the tables).

\begin{table}[h]
    \centering
    \caption{
    Summary of the experimental settings analysed in this study. `Continual training' refers to loading the model architecture and parameters as published, then training further on the same dataset as our model, under the same conditions (hyperparameters, prompts, etc.).
    Note that `CLIP+MLP-2' is equivalent to LLaVA-1.5 before any LLaVA-1.5 training.
    `Findings+' means we train on findings generation with GPT augmentation.
    }
    \footnotesize
    \setlength{\tabcolsep}{3pt}
    \centerline{%
    \begin{tabular}{@{}llllll@{}}
        \toprule
        Name & Description & Image encoder & Adapter (init.) & LLM (init.) &  Training\\
        \midrule
        LLaVA-1.0-init 
            & Continual training
            & CLIP-ViT-L-224px
            & Linear (LLaVA-1.0)
            & LLaMA-0 (LLaVA-1.0)
            & Findings \\
        LLaVA-Med-init 
            & Continual training
            & CLIP-ViT-L-224px
            & Linear (LLaVA-Med)
            & LLaMA-0 (LLaVA-Med)
            & Findings \\
        LLaVA-1.5-init 
            & Continual training
            & CLIP-ViT-L-336px
            & MLP-2 (LLaVA-1.5)
            & Vicuna-7B (LLaVA-1.5)
            & Findings \\
        \midrule
        CLIP+MLP-2 
            & General domain encoder
            & CLIP-ViT-L-336px
            & MLP-2 (Random)
            & Vicuna-7B
            & Findings \\
        \domainimageencoder+MLP-2 
            & +Use \domainimageencoder
            & \domainimageencoder-518px
            & MLP-2 (Random)
            & Vicuna-7B
            & Findings \\
        \domainimageencoder+MLP-4 
            & +Increase adapter size
            & \domainimageencoder-518px
            & MLP-4 (Random)
            & Vicuna-7B
            & Findings \\
        \maira{} 
            & +Use GPT-augmented data
            & \domainimageencoder-518px
            & MLP-4 (Random)
            & Vicuna-7B
            & Findings+ \\
       \bottomrule
    \end{tabular}%
    }
    \label{tab:describe_experiments}
\end{table}

\subsection{Adapting existing large multimodal models}
\maira{} training starts the multimodal alignment from scratch, fully relying on the findings generation task to align the image and text embeddings into a joint representation space.
To measure the effect of fine-tuning after aligning on other data, we compare to baselines fine-tuning three existing large multimodal models: LLaVA-1.0 \citep{liu_visual_2023} and LLaVA-1.5 \citep{liu_improved_2023} in the general domain, and LLaVA-Med \citep{li_llava-med_2023} in the biomedical domain.

\Cref{tab:baseline_results_test} (a) compares the performance of these settings.
We note that while fine-tuning LLaVA-1.5 outperforms training a comparable model from scratch (\Cref{tab:baseline_results_test}, `CLIP+MLP-2'), this advantage is lost when we replace CLIP with a domain-specific image encoder (\Cref{tab:baseline_results_test}, `\domainimageencoder{}+MLP-2'). Out of the different large multimodal models compared, we find that LLaVA-1.5 performs the best when fine-tuned for the findings generation task, although it lags behind \maira{}. 

Without such fine-tuning for findings generation, we observed that both LLaVA-1.0/1.5 and LLaVA-Med were incapable of producing meaningful radiology reports, using either the prompt from \citet{li_llava-med_2023} or our own (see qualitative examples in \Cref{fig:medpalm_example}).

\subsection{Which components are most beneficial?}
\maira{} differs from LLaVA-based multimodal models in its domain-specific image encoder, a deeper adapter module, and the use of a GPT-augmented dataset during training.
\Cref{tab:describe_experiments} describes several experiments showing the additive effect of these optimizations included in \maira{}, with results presented in \Cref{tab:baseline_results_test} (b). We start from an architecture mirroring LLaVA-1.5, using a pre-trained CLIP image encoder, a randomly initialized two-layer adapter, and Vicuna-7B as the \ac{LLM}. We train the adapter and \ac{LLM} jointly on in-domain radiology data to obtain an initial baseline (`CLIP+MLP-2'). Note that the performance of this model is slightly worse than that of a model fine-tuned from LLaVA-1.5, due to the random initialisation of the adapter.

By switching the image encoder from CLIP to a domain-specific model (\domainimageencoder) we observe an improvement across all metrics, more than compensating for the need to re-initialise the adapter.
The use of the domain-specific encoder also increases the number of image tokens from 576 to 1369. By correspondingly increasing the size of the adapter from two to four layers (`MLP-4'), we show further improvements across both clinical and lexical metrics.

As a final step, we augment the dataset with GPT-paraphrased samples (\Cref{sec:methods}) to produce \maira{}. The effect of the GPT-augmentation is to increase clinical metrics while slightly harming lexical metrics.
This could be explained by a mild distribution shift in the GPT-paraphrased reports respective to the original MIMIC reports.
In \Cref{sec:gpt_control} we demonstrate that gains from the use of GPT-paraphrased samples are not simply due to training longer.

\begin{table}
\centering
\caption{Effect of model design choices on findings generation performance. We report median and 95\% confidence intervals based on 500 bootstrap samples from the MIMIC-CXR test set. \textbf{(a)} Comparison of our approach to 3 baselines based on continually training LLaVA-style models \textbf{(b)} Additive gains from using \domainimageencoder, increasing the adapter size, and adding GPT-augmented data. \textbf{Bold} indicates best performance across all experiments. `$\downarrow$' indicates that lower is better (RadCliQ only).
CheXpert F1 metrics are computed based on CheXbert labeller ouputs.}
\footnotesize
\setlength{\tabcolsep}{5pt}
\centerline{%
\begin{tabular}{@{} l l|c c c|c c c|c @{}}\toprule
    \multicolumn{2}{@{}l}{\multirow{3}{*}{Metric}} & \multicolumn{3}{|c|}{\textbf{(a)} Continually trained baselines} & \multicolumn{3}{|c|}{\textbf{(b)} Additive optimizations} & \\
    \cmidrule(lr){3-5} \cmidrule(lr){6-8}
    &
        & \clap{\multirow{2}{*}{LLaVA-1.0-init}}
        & \clap{\multirow{2}{*}{LLaVA-Med-init}}
        & \clap{\multirow{2}{*}{LLaVA-1.5-init}}
        & CLIP
        & \domainimageencoder
        & \domainimageencoder
        & \maira{} \\
    & & & & & +MLP-2  & +MLP-2 & +MLP-4 & \\
    \midrule
    \multicolumn{2}{@{}l|}{Lexical:} &&&&&&& \\
    & ROUGE-L & 
        27.9 {\scriptsize\textcolor{gray}{[27.5, 28.5]}} & %
        27.6 {\scriptsize\textcolor{gray}{[27.1, 28.1]}} & %
        29.2 {\scriptsize\textcolor{gray}{[28.7, 29.7]}} & %
        28.2 {\scriptsize\textcolor{gray}{[27.8, 28.8]}} & %
        \bfseries 29.8 {\scriptsize\textcolor{gray}{[29.2, 30.3]}} & %
        \bfseries 30.1 {\scriptsize\textcolor{gray}{[29.6, 30.6]}} & %
        28.9 {\scriptsize\textcolor{gray}{[28.4, 29.4]}} \\
    & BLEU-1 & 
        35.5 {\scriptsize\textcolor{gray}{[34.8, 36.1]}} & %
        35.4 {\scriptsize\textcolor{gray}{[34.8, 36.0]}} & %
        35.6 {\scriptsize\textcolor{gray}{[35.0, 36.3]}} & %
        32.8 {\scriptsize\textcolor{gray}{[32.0, 33.5]}} & %
        35.4 {\scriptsize\textcolor{gray}{[34.7, 36.2]}} & %
        37.7 {\scriptsize\textcolor{gray}{[37.1, 38.4]}} & %
        \bfseries 39.2 {\scriptsize\textcolor{gray}{[38.7, 39.8]}} \\ %
    & BLEU-4 & 
        \bfseries 15.0 {\scriptsize\textcolor{gray}{[14.6, 15.5]}} & %
        \bfseries 14.9 {\scriptsize\textcolor{gray}{[14.5, 15.3]}} & %
        13.9 {\scriptsize\textcolor{gray}{[13.4, 14.3]}} & %
        12.7 {\scriptsize\textcolor{gray}{[12.2, 13.2]}} & %
        14.1 {\scriptsize\textcolor{gray}{[13.6, 14.6]}} & %
        \bfseries 14.9 {\scriptsize\textcolor{gray}{[14.4, 15.4]}} & %
        14.2 {\scriptsize\textcolor{gray}{[13.7, 14.7]}} \\ %
    & METEOR & 
        \bfseries 35.5 {\scriptsize\textcolor{gray}{[35.0, 35.9]}} & %
        \bfseries 35.3 {\scriptsize\textcolor{gray}{[34.8, 35.8]}} & %
        31.9 {\scriptsize\textcolor{gray}{[31.4, 32.5]}} & %
        30.3 {\scriptsize\textcolor{gray}{[29.8, 30.9]}} & %
        32.2 {\scriptsize\textcolor{gray}{[31.6, 32.7]}} & %
       33.4 {\scriptsize\textcolor{gray}{[32.8, 34.0]}} & %
       33.3 {\scriptsize\textcolor{gray}{[32.8, 33.8]}} \\ %
    \midrule
    \multicolumn{2}{@{}l|}{Clinical:} &&&&&&& \\
    & RadGraph-F1 & 
        19.9 {\scriptsize\textcolor{gray}{[19.3, 20.5]}} & %
        19.1 {\scriptsize\textcolor{gray}{[18.6, 19.7]}} & %
        21.5 {\scriptsize\textcolor{gray}{[20.9, 22.2]}} & %
        20.3 {\scriptsize\textcolor{gray}{[19.7, 20.9]}} & %
        23.0 {\scriptsize\textcolor{gray}{[22.4, 23.6]}} & %
        \bfseries 23.8 {\scriptsize\textcolor{gray}{[23.2, 24.5]}} & %
        \bfseries 24.3 {\scriptsize\textcolor{gray}{[23.7, 24.8]}}\\ %
    & RG\textsubscript{ER} & 
        24.4 {\scriptsize\textcolor{gray}{[23.8, 24.9]}} & %
        23.8 {\scriptsize\textcolor{gray}{[23.3, 24.4]}} & 
        26.4 {\scriptsize\textcolor{gray}{[25.9, 27.1]}} & %
        25.0 {\scriptsize\textcolor{gray}{[24.4, 25.6]}} &  %
        27.8 {\scriptsize\textcolor{gray}{[27.1, 28.4]}} & %
        28.8 {\scriptsize\textcolor{gray}{[28.3, 29.4]}} & %
        \bfseries 29.6 {\scriptsize\textcolor{gray}{[29.0, 30.2]}} \\ %
    & CheXbert vector & 
        37.7 {\scriptsize\textcolor{gray}{[36.8, 38.7]}} &  %
        36.9 {\scriptsize\textcolor{gray}{[36.0, 37.9]}} & %
        41.3 {\scriptsize\textcolor{gray}{[40.4, 42.2]}} & %
        39.6 {\scriptsize\textcolor{gray}{[38.7, 40.5]}} & %
        43.0 {\scriptsize\textcolor{gray}{[42.1, 43.9]}} & %
        \bfseries 43.8 {\scriptsize\textcolor{gray}{[43.0, 44.6]}} & %
        \bfseries 44.0 {\scriptsize\textcolor{gray}{[43.1, 44.9]}} \\ %
    & RadCliQ ($\downarrow$) & 
        3.27 {\scriptsize\textcolor{gray}{[3.23, 3.30]}} & %
        3.31 {\scriptsize\textcolor{gray}{[3.28, 3.35]}} & %
        3.22 {\scriptsize\textcolor{gray}{[3.18, 3.25]}} & %
        3.29 {\scriptsize\textcolor{gray}{[3.25, 3.32]}} & %
        \bfseries 3.14 {\scriptsize\textcolor{gray}{[3.10, 3.17]}} & %
        \bfseries 3.10 {\scriptsize\textcolor{gray}{[3.06, 3.13]}} & %
        \bfseries 3.10 {\scriptsize\textcolor{gray}{[3.07, 3.14]}}\\ %
    & \multicolumn{3}{l}{\it CheXpert F1, uncertain as negative:} &&&&& \\
    & \hspace{\tabcolsep} Macro-F1-14 & 
        25.5 {\scriptsize\textcolor{gray}{[24.2, 26.8]}} & %
        26.9 {\scriptsize\textcolor{gray}{[25.5, 28.5]}} & %
        29.6 {\scriptsize\textcolor{gray}{[28.3, 31.0]}} & %
        29.4 {\scriptsize\textcolor{gray}{[27.7, 30.8]}} & %
        33.4 {\scriptsize\textcolor{gray}{[31.8, 34.9]}} & %
        36.6 {\scriptsize\textcolor{gray}{[35.0, 38.3]}} & %
        \bfseries 38.6 {\scriptsize\textcolor{gray}{[37.1, 40.1]}} \\ %
    & \hspace{\tabcolsep} Micro-F1-14 & 
        43.6 {\scriptsize\textcolor{gray}{[42.3, 44.7]}} & %
        42.7 {\scriptsize\textcolor{gray}{[41.5, 44.0]}} & %
        49.0 {\scriptsize\textcolor{gray}{[47.8, 50.3]}} & %
        46.4 {\scriptsize\textcolor{gray}{[45.0, 47.6]}} & %
        52.2 {\scriptsize\textcolor{gray}{[51.1, 53.4]}} & %
        54.6 {\scriptsize\textcolor{gray}{[53.4, 55.8]}} & %
        \bfseries 55.7 {\scriptsize\textcolor{gray}{[54.7, 56.8]}} \\ %
    & \hspace{\tabcolsep} Macro-F1-5 & 
        36.5 {\scriptsize\textcolor{gray}{[34.7, 38.2]}} & %
        36.3 {\scriptsize\textcolor{gray}{[34.4, 38.1]}} & %
        41.7 {\scriptsize\textcolor{gray}{[39.8, 43.8]}} & %
        40.7 {\scriptsize\textcolor{gray}{[38.8, 42.6]}} & %
        43.2 {\scriptsize\textcolor{gray}{[41.7, 45.0]}} & %
        \bfseries 46.0 {\scriptsize\textcolor{gray}{[44.3, 48.1]}} & %
        \bfseries 47.7 {\scriptsize\textcolor{gray}{[45.6, 49.5]}} \\ %
    & \hspace{\tabcolsep} Micro-F1-5 & 
        45.2 {\scriptsize\textcolor{gray}{[43.4, 46.8]}} & %
        43.9 {\scriptsize\textcolor{gray}{[42.2, 45.6]}} & %
        50.5 {\scriptsize\textcolor{gray}{[48.9, 52.3]}} & %
        48.1 {\scriptsize\textcolor{gray}{[46.6, 49.8]}} & %
        53.6 {\scriptsize\textcolor{gray}{[52.2, 55.2]}} & %
        \bfseries 55.2 {\scriptsize\textcolor{gray}{[53.8, 56.8]}} & %
        \bfseries 56.0 {\scriptsize\textcolor{gray}{[54.5, 57.5]}} \\
    & \multicolumn{3}{l}{\it CheXpert F1, uncertain as positive:} &&&&& \\
    & \hspace{\tabcolsep} Macro-F1-14+ & 
        29.6 {\scriptsize\textcolor{gray}{[28.5, 30.6]}} & %
        30.6 {\scriptsize\textcolor{gray}{[29.3, 32.1]}} & %
        34.2 {\scriptsize\textcolor{gray}{[33.0, 35.5]}} & %
        33.0 {\scriptsize\textcolor{gray}{[31.6, 34.4]}} & %
        37.9 {\scriptsize\textcolor{gray}{[36.6, 39.3]}} & %
        40.4 {\scriptsize\textcolor{gray}{[38.9, 41.9]}} & %
        \bfseries 42.3 {\scriptsize\textcolor{gray}{[40.9, 43.6]}} \\ %
    & \hspace{\tabcolsep} Micro-F1-14+ & 
        44.5 {\scriptsize\textcolor{gray}{[43.4, 45.5]}} & %
        43.7 {\scriptsize\textcolor{gray}{[42.5, 44.8]}} & %
        49.6 {\scriptsize\textcolor{gray}{[48.4, 50.7]}} & %
        46.8 {\scriptsize\textcolor{gray}{[45.7, 48.0]}} & %
        52.6 {\scriptsize\textcolor{gray}{[51.5, 53.7]}} & %
        \bfseries 54.5 {\scriptsize\textcolor{gray}{[53.5, 55.6]}} & %
        \bfseries 55.3 {\scriptsize\textcolor{gray}{[54.3, 56.2]}} \\ %
    & \hspace{\tabcolsep} Macro-F1-5+ & 
        41.5 {\scriptsize\textcolor{gray}{[39.9, 43.0]}} &  %
        41.4 {\scriptsize\textcolor{gray}{[39.6, 43.1]}} &   %
        46.8 {\scriptsize\textcolor{gray}{[45.3, 48.6]}}  &   %
        45.6 {\scriptsize\textcolor{gray}{[43.9, 47.2]}} &   %
        49.4 {\scriptsize\textcolor{gray}{[47.8, 51.1]}}  &   %
        \bfseries 51.1 {\scriptsize\textcolor{gray}{[49.4, 52.8]}} &    %
        \bfseries 51.7 {\scriptsize\textcolor{gray}{[49.9, 53.1]}} \\  %
    & \hspace{\tabcolsep} Micro-F1-5+ & 
        48.5 {\scriptsize\textcolor{gray}{[47.0, 49.8]}} &   %
        47.1 {\scriptsize\textcolor{gray}{[45.4, 48.7]}} &   %
        54.0 {\scriptsize\textcolor{gray}{[52.4, 55.6]} } &   %
        51.5 {\scriptsize\textcolor{gray}{[50.1, 53.0]}} &   %
        56.7 {\scriptsize\textcolor{gray}{[55.4, 58.3]}}  &   %
        \bfseries 58.1 {\scriptsize\textcolor{gray}{[56.6, 59.5]}}  &   %
        \bfseries 58.8 {\scriptsize\textcolor{gray}{[57.4, 60.0]}}  \\  %
    \bottomrule
\end{tabular}%
}
\label{tab:baseline_results_test}
\end{table}

\subsection{How does \maira{} compare to existing approaches?}
\label{sec:sota_comparison}
Strict comparison with prior work is challenging due to variation in test set inclusion criteria and pre-processing steps, despite the existence of a `canonical' test split for MIMIC-CXR. For example, \cite{medpalm} starts from the official split, but includes lateral images paired with the study's report as independent samples, resulting in a reported test size of 4,834 images. \cite{evaluating_progress} and \cite{xrem} take only a single image for each study, resulting in 1,597 samples in the test set\footnote{We observed 2,210 studies after taking one image per study.}, and \cite{rgrg} follows the split provided by Chest ImaGenome~\citep{chest_imagenome}. Recall that our test set size is 2,461 image-report samples. We consider a full reproducibility study in the style of \cite{johnson2017reproducibility} out of scope for this work, however to enable future comparison we share the image identifiers used in our test set in an ancillary file. Changes in the distribution of the test set can significantly impact reported numbers, as demonstrated in \Cref{sec:stratified_results}. We attempt to account for some of this variability by reporting 95\% confidence intervals from bootstrap replicates on the test set when we compare with prior work, but numbers must be interpreted with caution. 

Given the above caveats, in \Cref{tab:SOTA_against_test} we compare \maira{} to prior work. For all lexical metrics, \maira{} seemingly outperforms or matches prior \ac{SOTA}. The substantial increase in METEOR relative to \cite{rgrg} is notable. On clinical metrics, there is no single superior approach---\cite{medpalm} report slightly superior scores on RadGraph-F1 while \cite{semantic_rewards} substantially outperforms on RG\textsubscript{ER}---this is perhaps expected given their model was optimised for RG\textsubscript{ER}. Across the CheXbert-derived metrics, \maira{} is comparable or superior across the fourteen-classes, but slightly underperforms on the five-class subset relative to \cite{medpalm}. We present class-stratified results for \maira{} in \Cref{sec:stratified_results} (\Cref{table:pathology_stratified}). Promisingly across all clinical metrics, \maira{} sets a new standard for the radiologist-aligned RadCliQ score.

\begin{table}
\centering
\caption{Findings generation performance on the MIMIC-CXR test set, compared to closest state-of-the-art for each metric. We report median with 95\% confidence intervals from 500 bootstrap samples of the test-set for \maira{}. Prior work uses differing test sets, limiting comparison. Our model is composed of a 86.6M parameter image encoder, 53M MLP adapter and 7B \ac{LLM}. Our test size is 2461 samples.  For the CheXpert F1 metrics, `+' means the uncertain class is mapped to positive, otherwise it is mapped to negative.
\textsuperscript{*}The CheXbert vector score is an evaluation of the \cite{miura} model performed by \cite{evaluating_progress}.
\textsuperscript{$\dagger$}The RadCliQ number is an evaluation of \cite{miura} model reported by \cite{xrem, evaluating_progress}, using RadCliQ-v0.}
\label{tab:SOTA_against_test}
\footnotesize
\begin{tabular}{@{} l l c | S@{ }l l r@{}}\toprule
    Category & Metric & \maira{} & {SOTA} & [ref.] & Param. count & Test set size\\
    & & & & & Vision / LLM &\\
    \midrule
    Lexical & 
    ROUGE-L & 
        \bfseries 28.9 {\scriptsize\textcolor{gray}{[28.4, 29.4]}} & 
        27.49 & \citep{medpalm} & 22B / 62B & 4,834 images\\
    & BLEU-1 &
        \bfseries 39.2 {\scriptsize\textcolor{gray}{[38.7, 39.8]}} &
        32.31 & \citep{medpalm} & 22B / 62B & 4,834 images \\
    & BLEU-4 & 
        \bfseries 14.2 {\scriptsize\textcolor{gray}{[13.7, 14.7]}} & 
        13.30 & \citep{miura} & 8M / ? & 2,347 reports\\
    & METEOR & 
        \bfseries 33.3 {\scriptsize\textcolor{gray}{[32.8, 33.8]}} & 
        16.8 & \citep{rgrg} & 26M / 355M & 32,711 images\\
    \midrule
    Clinical & 
    RadGraph-F1 & 
        24.3 {\scriptsize\textcolor{gray}{[23.7, 24.8]}} &
        \bfseries 26.71 & \citep{medpalm} & 22B / 62B & 4,834 images\\
    & RG\textsubscript{ER} & 
        29.6 {\scriptsize\textcolor{gray}{[29.0, 30.2]}} & 
        \bfseries 34.7 & \citep{semantic_rewards} & 8M / 18M & 2,347 reports\\  %
    & CheXbert vector & 
        44.0 {\scriptsize\textcolor{gray}{[43.1, 44.9]}} &  %
        \bfseries 45.2 & 
        \citep{miura}* & 
        8M / ? &
        1,597 images\\
    & RadCliQ ($\downarrow$) & 
        \bfseries 3.10 {\scriptsize\textcolor{gray}{[3.07, 3.14]}} & 
        {3.277} & 
        \citep{miura}$^\dagger$ & 
        8M / ? & 1,597 images\\
    \cmidrule{2-7}
    & Macro-F1-14 & 
        \bfseries 38.6 {\scriptsize\textcolor{gray}{[37.1, 40.1]}} &
        \bfseries 39.83 & \citep{medpalm} & 22B / 62B & 4,834 images \\
    & Micro-F1-14 & 
        \bfseries 55.7 {\scriptsize\textcolor{gray}{[54.7, 56.8]}} &
        53.56 & \citep{medpalm} & 22B / 62B & 4,834 images \\
    & Macro-F1-5 & 
        47.7 {\scriptsize\textcolor{gray}{[45.6, 49.5]}} & 
        \bfseries 51.60 & \citep{medpalm} & 22B / 62B & 4,834 images \\
    & Micro-F1-5 & 
        56.0 {\scriptsize\textcolor{gray}{[54.5, 57.5]}} & 
        \bfseries 57.88 & \citep{medpalm} & 22B / 62B & 4,834 images \\
    \cmidrule{2-7}
    & Macro-F1-14+ & 
        42.3 {\scriptsize\textcolor{gray}{[40.9, 43.6]}} &
        {--} & \\
    & Micro-F1-14+ & 
        55.3 {\scriptsize\textcolor{gray}{[54.3, 56.2]}} & 
        {--} & \\
    & Macro-F1-5+ & 
        51.7 {\scriptsize\textcolor{gray}{[49.9, 53.1]}} &
        {--} & \\
    & Micro-F1-5+ & 
        \bfseries 58.8 {\scriptsize\textcolor{gray}{[57.4, 60.0]}} &
        54.7 & \citep{rgrg} & 26M / 355M & 32,711 images\\
    \bottomrule
\end{tabular}
\end{table}

\subsection{Stratified results}
\label{sec:stratified_results}
\paragraph{Performance depends on finding class}

\begin{table}[h]
\centering
\caption{Breakdown of metrics per CheXpert finding class, as defined by \cite{chexpert}. Classes are hierarchical and not mutually exclusive. The positive class is `present', and `uncertain' is mapped to negative. `Lung Lesion' includes masses, nodular densities  or opacities, lumps, and tumors. `Pleural Other' includes pleural or parenchymal thickening or scarring, as well as fibrosis. `Support Devices' includes lines, tubes, catheters, pacemakers, coils, drains, etc. NPV = negative predictive value. Performance numbers are median and 95\% confidence intervals from 500 bootstrap replicates from the MIMIC-CXR test set.}
\label{table:pathology_stratified}
\footnotesize
\centerline{%
\begin{tabular}{@{} lr@{ }l|lllll @{}}
\toprule
 Finding  class & \multicolumn{2}{c}{\hspace{-5pt}\% ($n$, median)}  & Precision & Recall & NPV & Specificity & F$_1$-score \\
\midrule
    No Finding & 
        6\% & (151) %
        & 31.6 {\scriptsize\textcolor{gray}{[26.3, 37.9]}}
        & 49.1 {\scriptsize\textcolor{gray}{[41.7, 56.8]}}
        & 96.6 {\scriptsize\textcolor{gray}{[95.8, 97.3]}}
        & 93.1 {\scriptsize\textcolor{gray}{[92.1, 94.0]}}
        & 38.6 {\scriptsize\textcolor{gray}{[32.6, 44.5]}}\\\midrule
    Lung Opacity & 
        38\% & (944) %
        & 58.0 {\scriptsize\textcolor{gray}{[54.6, 61.6]}}
        & 43.7 {\scriptsize\textcolor{gray}{[40.5, 46.6]}}
        & 69.6 {\scriptsize\textcolor{gray}{[67.7, 71.9]}}
        & 80.3 {\scriptsize\textcolor{gray}{[78.2, 82.2]}}
        & 49.8 {\scriptsize\textcolor{gray}{[47.1, 52.7]}}\\
    Atelectasis & 
        28\% & (688) %
        & 43.3 {\scriptsize\textcolor{gray}{[39.5, 47.3]}}
        & 39.4 {\scriptsize\textcolor{gray}{[35.6, 43.4]}}
        & 77.3 {\scriptsize\textcolor{gray}{[75.4, 79.1]}}
        & 79.9 {\scriptsize\textcolor{gray}{[78.0, 81.8]}}
        & 41.3 {\scriptsize\textcolor{gray}{[37.8, 44.6]}}\\
    Edema & 
        18\% & (436) %
        & 47.4 {\scriptsize\textcolor{gray}{[42.1, 52.6]}}
        & 41.2 {\scriptsize\textcolor{gray}{[37.1, 45.9]}}
        & 87.7 {\scriptsize\textcolor{gray}{[86.4, 89.0]}}
        & 90.1 {\scriptsize\textcolor{gray}{[88.6, 91.4]}}
        & 44.0 {\scriptsize\textcolor{gray}{[39.9, 48.6]}}\\
    Lung Lesion 
        & 6\% & (146) %
        & 30.1 {\scriptsize\textcolor{gray}{[18.4, 41.6]}}
        & 13.6 {\scriptsize\textcolor{gray}{[7.9, 18.6]}}
        & 94.7 {\scriptsize\textcolor{gray}{[93.9, 95.5]}}
        & 98.0 {\scriptsize\textcolor{gray}{[97.4, 98.5]}}
        & 18.8 {\scriptsize\textcolor{gray}{[11.4, 24.9]}}\\
    Consolidation 
        & 5\% & (115) %
        & 25.9 {\scriptsize\textcolor{gray}{[16.3, 36.2]}}
        & 16.4 {\scriptsize\textcolor{gray}{[10.2, 23.8]}}
        & 96.0 {\scriptsize\textcolor{gray}{[95.1, 96.8]}}
        & 97.7 {\scriptsize\textcolor{gray}{[97.0, 98.3]}}
        & 20.0 {\scriptsize\textcolor{gray}{[12.9, 28.1]}}\\
    Pneumonia 
        & 5\% & (113) %
        & 22.1 {\scriptsize\textcolor{gray}{[13.7, 32.4]}}
        & 15.5 {\scriptsize\textcolor{gray}{[9.6, 22.8]}}
        & 96.0 {\scriptsize\textcolor{gray}{[95.2, 96.9]}}
        & 97.3 {\scriptsize\textcolor{gray}{[96.7, 98.0]}}
        & 18.3 {\scriptsize\textcolor{gray}{[11.7, 25.5]}}\\\midrule
    Cardiomegaly 
        & 37\% & (903) %
        & 61.7 {\scriptsize\textcolor{gray}{[58.7, 64.5]}}
        & 66.3 {\scriptsize\textcolor{gray}{[63.2, 69.5]}}
        & 79.6 {\scriptsize\textcolor{gray}{[77.6, 81.8]}}
        & 76.2 {\scriptsize\textcolor{gray}{[74.1, 78.1]}}
        & 64.0 {\scriptsize\textcolor{gray}{[61.2, 66.6]}}\\
    Enlarged Cardiomediastinum 
        & 8\% & (196)%
        & 13.2 {\scriptsize\textcolor{gray}{[8.1, 18.6]}}
        & 10.6 {\scriptsize\textcolor{gray}{[6.8, 15.2]}}
        & 92.4 {\scriptsize\textcolor{gray}{[91.4, 93.5]}}
        & 93.9 {\scriptsize\textcolor{gray}{[92.9, 94.9]}}
        & 11.9 {\scriptsize\textcolor{gray}{[7.4, 16.4]}}\\\midrule
    Pleural Effusion 
        & 34\% & (833) %
        & 69.0 {\scriptsize\textcolor{gray}{[65.9, 72.2]}}
        & 68.6 {\scriptsize\textcolor{gray}{[65.3, 72.0]}}
        & 83.9 {\scriptsize\textcolor{gray}{[82.1, 86.0]}}
        & 84.3 {\scriptsize\textcolor{gray}{[82.3, 85.9]}}
        & 68.9 {\scriptsize\textcolor{gray}{[66.3, 71.3]}}\\
    Pleural Other 
        & 3\% & (77) %
        & 23.7 {\scriptsize\textcolor{gray}{[10.1, 38.4]}}
        & 10.8 {\scriptsize\textcolor{gray}{[4.3, 18.2]}}
        & 97.2 {\scriptsize\textcolor{gray}{[96.6, 97.8]}}
        & 98.9 {\scriptsize\textcolor{gray}{[98.5, 99.3]}}
        & 14.7 {\scriptsize\textcolor{gray}{[6.2, 23.5]}}\\
    Pneumothorax 
        & 2\% & (55) %
        & 34.2 {\scriptsize\textcolor{gray}{[24.4, 44.4]}}
        & 50.8 {\scriptsize\textcolor{gray}{[37.4, 65.5]}}
        & 98.9 {\scriptsize\textcolor{gray}{[98.4, 99.3]}}
        & 97.8 {\scriptsize\textcolor{gray}{[97.2, 98.3]}}
        & 40.8 {\scriptsize\textcolor{gray}{[29.7, 51.2]}}\\\midrule
    Fracture 
        & 5\% & (130) %
        & 37.0 {\scriptsize\textcolor{gray}{[25.9, 50.0]}}
        & 18.8 {\scriptsize\textcolor{gray}{[12.3, 25.8]}}
        & 95.6 {\scriptsize\textcolor{gray}{[94.8, 96.4]}}
        & 98.2 {\scriptsize\textcolor{gray}{[97.7, 98.7]}}
        & 24.9 {\scriptsize\textcolor{gray}{[17.3, 33.1]}}\\\midrule
    Support Devices 
        & 41\% & (1001) %
        & 84.6 {\scriptsize\textcolor{gray}{[82.4, 86.7]}}
        & 84.4 {\scriptsize\textcolor{gray}{[82.0, 86.4]}}
        & 89.3 {\scriptsize\textcolor{gray}{[87.7, 90.8]}}
        & 89.4 {\scriptsize\textcolor{gray}{[87.9, 91.1]}}
        & 84.5 {\scriptsize\textcolor{gray}{[82.7, 86.0]}}\\
\bottomrule
\end{tabular}%
}
\end{table}

\Cref{table:pathology_stratified} shows a breakdown of \maira{} performance based on the finding class. We use the standard 14 hierarchical classes initially proposed by \cite{chexpert} and used in the CheXbert labeller~\citep{chexbert}.

Looking at F$_1$-score, we note that the Macro-F1-14 values we report elsewhere mask variance across classes. We see high-performing classes (Support Devices: 84.5, Pleural Effusion: 68.9, Cardiomegaly: 64.0) and indeed poorly performing classes (Enlarged Cardiomediastinum: 11.9, Pleural Other: 14.7, Pneumonia: 18.3). We note that these latter categories are rarer and more nebulously defined,\footnote{As acknowledged by \cite{chexpert}, pneumonia is a clinical diagnosis which should strictly not be assessed from a chest X-ray alone.} and may be subject to more noise in the output of the CheXbert labeller itself. For example, Cardiomegaly (37\% prevalence) should imply an enlarged cardiomediastinum, and yet the latter category occurs in only 8\% of cases. %

Specificity and \ac{NPV} are not measured in the Macro-F1-14 aggregate metric, where we see consistently high values. Compared to recall and precision (positive predictive value), we infer the model may under- or miss-call positive findings, but more reliably reports on the absence of findings. This is useful for findings such as pneumothorax (NPV 98.9) in an intensive care setting, where a chest X-ray may serve to confirm the \emph{absence} of a pneumothorax after the insertion of a drain or tube.

\paragraph{Results differ between normal and abnormal studies}
When there are no findings in a study, radiology reports are often formulaic, containing templated phrases such as ``No [evidence of] acute cardiopulmonary process".
This constitutes a simpler language generation task for the model, as reflected in the higher lexical metrics in \Cref{tab:stratified_results:normal_vs_abnormal_results}.
In addition, we observe higher clinical metrics in the no-finding subset, which we speculate could be related to the distribution shift between MIMIC-CXR dataset splits.
As noted by \citet{johnson2019mimic-cxr-jpg}, studies with no findings are over-represented in the released training and validation splits, compared to the test set.

\paragraph{The model benefits from the indication section}
\begin{table}
\centering
\caption{Breakdown of \maira{} metrics by (i) whether the study has the CheXpert label `No finding' and (ii) whether the report contains an indication for the study.
(i) For `normal' cases (no finding), the original reports tend to be more formulaic---which could explain the higher lexical metrics in this group---as well as defining a somewhat simpler clinical task for the model.
(ii) The results are strikingly superior for cases that contain an indication, as the model is able to leverage the strong cues for what findings (positive or negative) should be reported. All performance numbers are median and 95\% confidence intervals from 500 bootstrap replicates from the MIMIC-CXR test set.}
\label{tab:stratified_results:normal_vs_abnormal_results}
\footnotesize
\begin{tabular}{@{} ll|cc|cc @{}}\toprule
    Category & Metric & Has finding & No finding & Has indication & No indication \\
    \midrule
    & \% ($n$) &
        78.3\% (1928)&
        21.7\% (533)&
        57.5\% (1414)&
        42.5\% (1047) \\
    \midrule
    Lexical 
    & ROUGE-L & 
        27.6 {\scriptsize \textcolor{gray}{ [27.1, 28.1]}} & 
        \bfseries 33.4 {\scriptsize \textcolor{gray}{ [31.9, 34.9]}} & 
        \bfseries 32.7 {\scriptsize \textcolor{gray}{ [32.0, 33.4]}} & 
        23.6 {\scriptsize \textcolor{gray}{ [23.1, 24.1]}} \\
    & BLEU-4 & 
        13.2 {\scriptsize \textcolor{gray}{ [12.7, 13.7]}} & 
        \bfseries 18.7 {\scriptsize \textcolor{gray}{ [17.4, 20.3]}} & 
        \bfseries 17.6 {\scriptsize \textcolor{gray}{ [16.9, 18.2]}} & 
        9.0 {\scriptsize \textcolor{gray}{ [8.6, 9.6]}} \\
    & METEOR & 
        31.9 {\scriptsize \textcolor{gray}{ [31.4, 32.5]}} & 
        \bfseries 38.5 {\scriptsize \textcolor{gray}{ [37.1, 40.0]}} & 
        \bfseries 36.8 {\scriptsize \textcolor{gray}{ [36.1, 37.6]}} & 
        28.6 {\scriptsize \textcolor{gray}{ [28.0, 29.3]}} \\    
    \midrule
    Clinical & 
    RadGraph-F1 & 
        23.0 {\scriptsize \textcolor{gray}{ [22.4, 23.5]}} & 
        \bfseries 28.5 {\scriptsize \textcolor{gray}{ [27.0, 29.9]}} & 
        \bfseries 27.8 {\scriptsize \textcolor{gray}{ [27.0, 28.5]}} & 
        19.2 {\scriptsize \textcolor{gray}{ [18.6, 20.0]}} \\
    & 
    RG\textsubscript{ER} & 
        28.5 {\scriptsize \textcolor{gray}{ [27.9, 29.0]}} & 
        \bfseries 33.7 {\scriptsize \textcolor{gray}{ [32.0, 35.3]}}  & 
        \bfseries 33.5 {\scriptsize \textcolor{gray}{ [32.8, 34.3]}}  & 
        24.3 {\scriptsize \textcolor{gray}{ [23.6, 25.0]}} \\
    & ChexBert vector & 
        42.3 {\scriptsize \textcolor{gray}{ [41.3, 43.2]}} & 
        \bfseries 50.3 {\scriptsize \textcolor{gray}{ [48.1, 52.2]}} & 
        \bfseries 47.3 {\scriptsize \textcolor{gray}{ [46.2, 48.3]}} & 
        39.5 {\scriptsize \textcolor{gray}{ [38.2, 40.8]}}\\
    & RadCliQ ($\downarrow$) &
        3.19 {\scriptsize \textcolor{gray}{ [3.16, 3.22]}} & 
        \bfseries 2.79 {\scriptsize \textcolor{gray}{ [2.71, 2.88]}} &
        \bfseries 2.88 {\scriptsize \textcolor{gray}{ [2.84, 2.92]}} & 
        3.41 {\scriptsize \textcolor{gray}{ [3.37, 3.45]}}\\
    \cmidrule{2-6}
    & Macro-F1-14 & 
        38.1 {\scriptsize \textcolor{gray}{ [36.2, 39.9]}} & 
        -- & 
        39.1 {\scriptsize \textcolor{gray}{ [37.1, 41.5]}} & 
        36.9 {\scriptsize \textcolor{gray}{ [34.4, 39.2]}} \\
    & Micro-F1-14 & 
        57.0 {\scriptsize \textcolor{gray}{ [55.8, 57.9]}} & 
        -- & 
        56.3 {\scriptsize \textcolor{gray}{ [54.7, 57.8]}} & 
        55.0 {\scriptsize \textcolor{gray}{ [53.3, 56.6]}} \\
    & Macro-F1-5  & 
        48.8 {\scriptsize \textcolor{gray}{ [46.7, 50.8]}} & 
        -- & 
        46.8 {\scriptsize \textcolor{gray}{ [44.5, 50.0]}} & 
        47.3 {\scriptsize \textcolor{gray}{ [44.4, 50.4]}} \\
    & Micro-F1-5 & 
        57.5 {\scriptsize \textcolor{gray}{ [56.0, 59.0]}} & 
        -- & 
        56.6 {\scriptsize \textcolor{gray}{ [54.6, 58.6]}} & 
        55.5 {\scriptsize \textcolor{gray}{ [53.3, 57.8]}} \\
    \bottomrule
\end{tabular}
\end{table}

The study indication is expected to substantially affect the contents of a radiology report. For example, as in \Cref{tab:gpt_example}, it may prompt the radiologist to report on specific types of abnormality that might not be routinely included, and is particularly relevant for deciding what to report as absent from the scan.
\maira{} uses the indication section of the report to help generate the findings section, whenever it is available with the study (66.3\% in training and 57.5\% in the test set).
In  \Cref{tab:stratified_results:normal_vs_abnormal_results} we show how performance varies in subsets of the test set with and without the indication section.
Indeed, by leveraging the indication, \maira{} is able to to generate radiology reports that are drastically more similar and more accurate with respect to the true report than in the subset without indication.

    \section{Examples}

\newcommand{\mairaexample}[4]{%
    \begin{minipage}[c]{0.35\textwidth}
        \vspace{0cm}
        \includegraphics[width=\linewidth]{#1}
    \end{minipage}
    \hfill
    \begin{minipage}[c]{0.63\textwidth}
        \begin{tabularx}{\linewidth}{@{}lX@{}}
            \toprule
            Original report & #2 \\ & #3 \\
            \midrule
            \maira{} & #4 \\
            \bottomrule
        \end{tabularx}
    \end{minipage}%
}

\begin{figure}
\centering
\begin{minipage}[c]{0.4\textwidth}
    \vspace{0cm} %
    \includegraphics[scale=0.3]{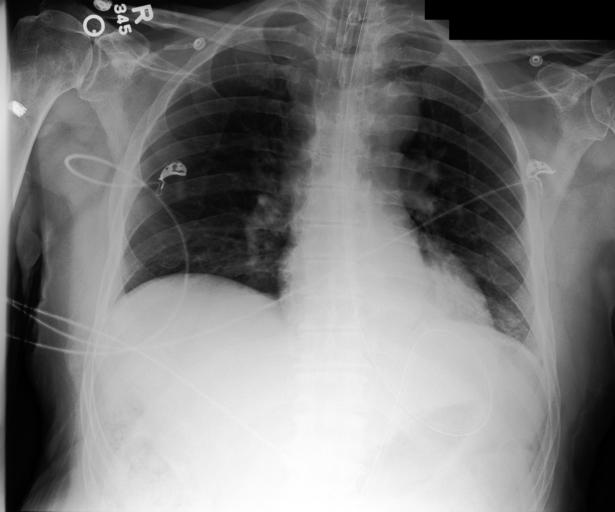}
\end{minipage}
\hfill
\begin{minipage}[c]{0.58\textwidth}
    \vspace{0cm} %
    \begin{tabularx}{\linewidth}{@{}X@{}}
        \toprule
        Original report \\
        \midrule
        INDICATION: \_-year-old male with intubation. No prior examinations for comparison. \\[6mm]
        FINDINGS: Endotracheal tube terminates 6 cm from the carina. Nasogastric tube loops within the stomach, with side port within the gastric body. There are no significant pleural effusions or pneumothorax. Mild retrocardiac atelectasis is present. Heart size is normal. The aorta is tortuous. \\
        \bottomrule
    \end{tabularx}
\end{minipage}

\vspace*{0.5cm}
\begin{tabularx}{\textwidth}{@{}lX@{}}\toprule
    Model & Generated findings \\
    \midrule
    \maira{} &
    Endotracheal tube terminates 5 cm above the carina. Nasogastric tube loops in the stomach, with \reporterror{side port near the GE junction}. Heart size is normal. Aorta is tortuous and unfolded. There are no pleural effusions or pneumothorax. \reporterror{Lungs are clear.} \reportomission{No comment about atelectasis.} \\
    \midrule
    Med-PaLM  M-12B &
    An endotracheal tube terminates 5 cm above the carina. There is no pneumothorax or large pleural effusion. Opacities in the \reporterror{medial right lower lobe} and retrocardiac region may represent atelectasis aspiration or pneumonia. \reporterror{A calcified lymph node is noted in the aortopulmonary window.} Cardiac size is normal. \reportomission{No comment about NG/OG tube.}\\
    \midrule
    Med-PaLM M-84B &
    Endotracheal tube terminates 5 cm above the carina. Nasogastric tube courses \reporterror{below the diaphragm and inferiorly beyond the film}. There is retrocardiac opacity which may represent atelectasis or aspiration. The heart size is normal. The aorta is tortuous and unfolded. No pleural effusions or pneumothorax. \\
    \midrule
    LLaVA-Med &
    \textcolor{gray}{The image is a chest X-ray taken on day 1 post-admission. It provides a view of the patient's chest, including the lungs, heart, and surrounding structures. Chest X-rays are commonly used to help diagnose and monitor various conditions affecting the respiratory and cardiovascular systems, such as infections, fluid accumulation, or structural abnormalities.} \\
    \midrule
    LLaVA-1.5 &
    \textcolor{gray}{The image displays a close-up view of a human body, focusing on the chest area. The chest is filled with various medical instruments, including a heart monitor and a defibrillator. The heart monitor is placed on the left side of the chest, while the defibrillator is positioned on the right side.} 
    
    \textcolor{gray}{In addition to the medical instruments, there are several wires and cords scattered throughout the scene. Some of these wires are attached to the medical devices, while others are likely part of the electrical connections for the instruments. The presence of these wires and cords suggests that the medical equipment is in use, possibly for monitoring or treating a patient.}\\
    \bottomrule
\end{tabularx}
\caption{Model outputs on the sample from \cite{medpalm}. \maira{} and Med-PaLM M-84B produce similar reports with different errors; \maira{} misses the retrocardiac atelectasis and incorrectly locates the side port of the nasogastric tube, whereas Med-PaLM M-84B incorrectly describes the NG tube as ending beyond the film. For comparison purposes, we include Med-PaLM M-12B, which hallucinates a calcified lymph node and fails  to mention the NG tube, and the outputs from  LLaVA-Med and LLaVA-1.5, which do not produce realistic radiology reports.}
\label{fig:medpalm_example}
\end{figure}

\Cref{fig:medpalm_example} reproduces the example%
    \footnote{In this section we show the JPEG versions of images as released in MIMIC-CXR-JPG \citep{johnson2019mimic-cxr-jpg}. However, as described in \Cref{sec:methods}, \maira{} ingests images derived from the original DICOMs.}
shown in \cite{medpalm}, comparing the output of \maira{} with the best Med-PaLM M variant (84B) and the closest by model size (12B). For illustration, we also include the output of models not trained on the findings generation task (LLaVA-1.5 and LLaVA-Med).

\begin{figure}

\centering
\mairaexample{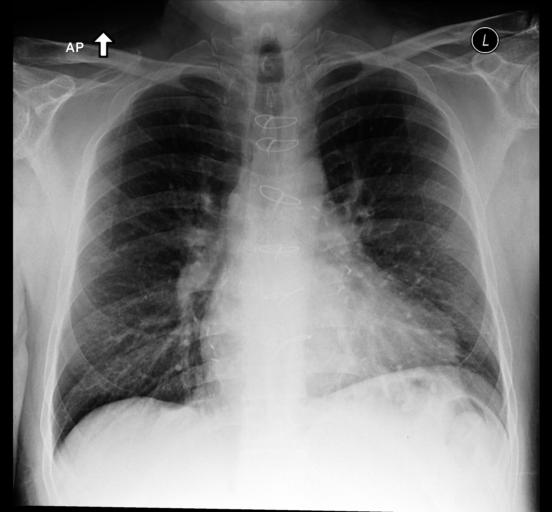}{
INDICATION: \_-year-old man with cough and fever, question acute process.
}{
FINDINGS: The lungs are well expanded and clear. The cardiomediastinal silhouette, hilar contours, and pleural surfaces are normal. No pleural effusion or pneumothorax is present. \reporthighlight{Sternal wires are intact.}
}{
FINDINGS: The lungs are well expanded and clear. The cardiomediastinal silhouette, hilar contours, and pleural surfaces are normal. No pleural effusion or pneumothorax is present. \reporthighlight{Median sternotomy wires and surgical clips are noted.}
}
\caption{\maira{} successfully generates the \findings{} section for a normal chest X-ray. \reporthighlight{Highlighted} is a small variation in how the sternal wires are described; here the original report is slightly more specific, but \maira{} is not incorrect. RadCliQ: 0.239.}
\label{fig:boring_example}
\end{figure}

To conduct further error analysis, we sorted all examples in the test set by their RadCliQ score and manually reviewed the top and bottom cases with two consultant radiologists. The highest-scoring examples (lowest RadCliQ) were largely cases without findings, where the report follows a standard format; this is seen already in \Cref{tab:stratified_results:normal_vs_abnormal_results} when we stratify by whether the case has a finding. \Cref{fig:boring_example} shows such an example; here \maira{} has generated an almost-identical report to that of the original radiologist, with the exception of providing less detail on the sternal wires. 

A recurring but expected `failure' mode we observe in \maira{} is the generation of descriptions of change. This is a well-known problem in the generation of radiology reports \citep{bannur2023learning,ramesh2022improving}: Standard clinical practice involves reporting using both current and prior studies precisely to enable assessment of change. This version of \maira{} is trained on reports from single images, where reports often contain descriptions of changes compared to a previous study. This type of report is then also sometimes produced at test time, as illustrated in some of the following examples.

\begin{figure}

\centering
\mairaexample{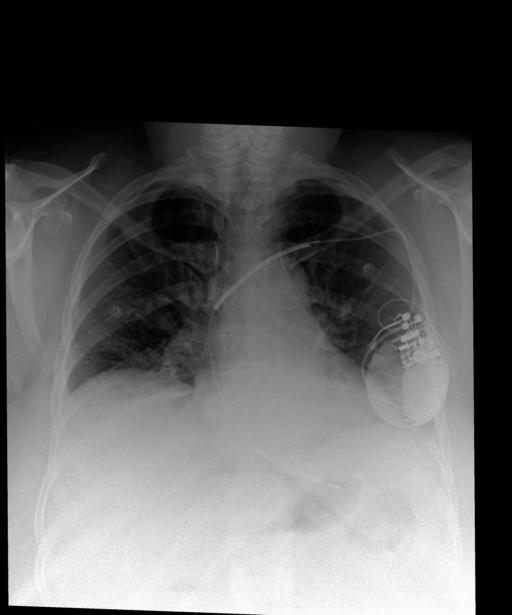}{
INDICATION: \_F with shortness of breath, congestive heart failure.
}{
FINDINGS: \reporthighlight{Lung volumes are low compared to the previous study.} Left-sided AICD device is noted with single lead terminating in unchanged position in the right ventricle. Heart size appears at least mildly enlarged. The mediastinal and hilar contours are unremarkable. There is crowding of the bronchovascular structures without overt pulmonary edema. Streaky opacities in the lung bases likely reflect areas of atelectasis. No pleural effusion or pneumothorax is present. There are no acute osseous abnormalities.
}{
FINDINGS: \reporthighlight{Lung volumes are low.} Left-sided AICD device is noted with single lead terminating in the region of the right ventricle. Heart size is mildly enlarged. Mediastinal and hilar contours are unremarkable. Crowding of bronchovascular structures is present without overt pulmonary edema. Patchy opacities in the lung bases likely reflect areas of atelectasis. No pleural effusion or pneumothorax is present. There are no acute osseous abnormalities.
}
\caption{\maira{} generates the \findings{} section for a more complex case with multiple findings. Here, the original report describes the lung volumes in relation to a prior image which is not available to \maira{}; in this case \maira{} simply describes the lung volumes as low (\reporthighlight{highlighted}). RadCliQ: 0.239.}
\label{fig:success_example1}
\end{figure}

\begin{figure}

\centering
\mairaexample{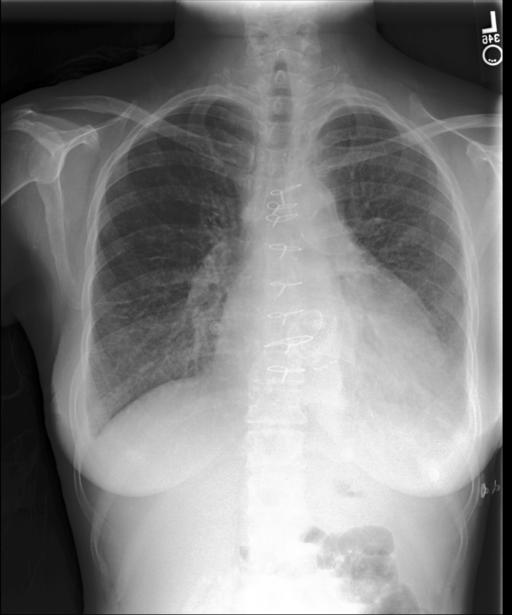}{
INDICATION: \emph{[No indication]}
}{
FINDINGS: The patient is status post median sternotomy, CABG, and mitral valve replacement. The heart is mildly enlarged. The mediastinal contours are unchanged with calcification of the aortic knob again noted. Mild pulmonary edema appears progressed compared to the prior exam with small bilateral pleural effusions, also minimally increased compared to the prior exam. Left basilar opacification likely reflects atelectasis. There is no pneumothorax. No acute osseous abnormalities are identified.
}{
FINDINGS: The patient is status post median sternotomy and mitral valve replacement. Heart size is moderately enlarged \reporterror{but unchanged}. The mediastinal contours are stable. There is mild interstitial pulmonary edema, \reporterror{new compared to the prior study}. Small bilateral pleural effusions are \reporterror{also new}. Retrocardiac opacity likely reflects atelectasis. There is no pneumothorax. No acute osseous abnormalities are visualized.
}
\caption{\maira{} generates the \findings{} section of a complex case. Here, although \maira{} correctly describes the presence and absence of findings, it generates descriptions of change which cannot be justified from a single image. RadCliQ: 1.053.}
\label{fig:success_example2}
\end{figure}

\Cref{fig:success_example1,fig:success_example2} show success cases from \maira{}. In \Cref{fig:success_example1}, \maira{} successfully \emph{avoids} hallucinating a comparison to a prior study  (`Lung volumes are low.'), while \Cref{fig:success_example2} shows examples of such hallucination (`... pulmonary edema, new compared to the prior study.'), with other details largely correct.

\begin{figure}

\centering
\mairaexample{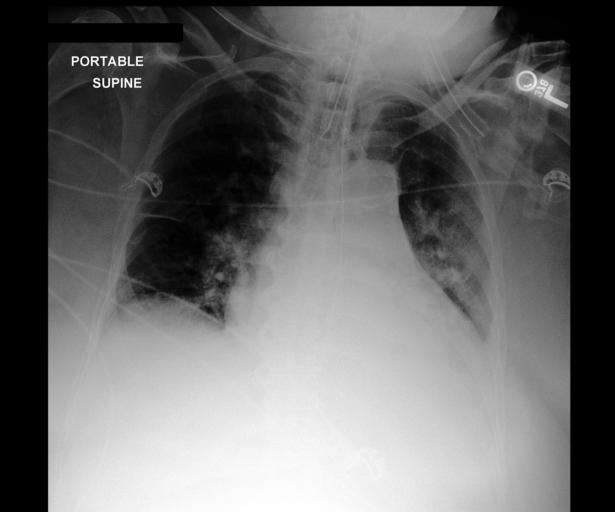}{
INDICATION: \emph{[No indication]}
}{
FINDINGS: \reporthighlight{Compared to the prior exam there is no significant interval change.}
}{
FINDINGS: The endotracheal tube is positioned 4 cm above the carina, while the nasogastric tube is located below the diaphragm. The cardiac silhouette is enlarged with pulmonary vascular congestion. The left base shows opacification, indicating volume loss in the lower lobe and pleural effusion. The right base shows less prominent atelectatic changes and a possible small effusion.
}
\caption{The original \findings{} section in this report does not describe any of the observed findings, comparing only to the previous image. Here, \maira{} correctly describes the findings in the image. The RadCliQ score (5.327) is high due to low overlap between the reports, unfairly penalising the model's output.}
\label{fig:middle_example}
\end{figure}

\Cref{fig:middle_example} shows a case with high (poor) RadCliQ score, however this does not reflect a failure of the model and rather a limitation of the evaluation. In this case, the original report contains negligible detail (`Compared to the prior exam there is no significant interval change.'), whereas \maira{} generates a full report which is largely correct. We note that examples of the converse also exist, where \maira{} simply refers to the absence of change without further elaboration (for 491 studies, the \findings{} section is exactly the sentence `Compared to the prior study there is no significant interval change.'). This underscores a limitation of training with and evaluating on `noisy' real-world datasets.%

We further note that, in \Cref{fig:middle_example}, \maira{} has generated a quantitative measurement (`... tube is positioned 4cm above the carina'), which is not grounded in physical measurements of the image. This is also observed in \Cref{fig:medpalm_example} from both \maira{} and Med-PaLM M.
Whereas a model may learn about an average field-of-view of the images seen during training, as well as certain correlations in the training reports, such measurements cannot be produced accurately without knowledge of physical and geometric parameters of the image acquisition.

    \section{Discussion}
We have presented results on \maira{}, a radiology-specialised large multimodal model designed for the task of generating the \findings{} section of a chest X-ray report.

Architecturally, \maira{} consists of a frozen domain-specific image encoder based on a ViT-B (\domainimageencoder), a four-layer feedforward adapter module, and the \ac{LLM} Vicuna-7B. We train \maira{} solely on the open-access MIMIC-CXR dataset, leveraging GPT-3.5 for text-based data augmentation.
With these components, we demonstrate that performance competitive with existing state-of-the-art is possible, with either fewer parameters or a simpler training objective. %

The use of a domain-specific image encoder significantly boosts the performance of \maira{}. One aspect of this may be the increased image resolution of \domainimageencoder{} (518px), enabling the detection of potentially small and/or subtle findings such as pneumothorax. A larger set of image tokens enabled us to fruitfully scale up the adapter layer, raising the prospect of further gains from yet more complex processing of image tokens.

By using GPT-3.5 to paraphrase reports in MIMIC-CXR, we observe a boost in clinical metrics with a minor degradation of lexical metrics. We speculate that this paraphrasing serves as a semantics-preserving transformation of the text, encouraging the model to focus on the key aspects of the report without overfitting to its style.

Our stratified analysis reveals significantly higher performance on studies containing an \indication{} section. We hypothesise two mechanisms for this. Firstly, knowledge of the `question' behind the report may inform the expected \findings{}; for example,  `...please evaluate for pleural effusion' (\indication{}) prompts: `Left mid to lower lung opacified is likely a combination of moderate right-sided pleural effusion...' (\findings{}).
Secondly, the \indication{} section can include additional clinical context on the patient, which has been shown to improve interpretation~\citep{yapp2022effect}.

By stratifying by findings class, we observe that the behaviour of \maira{} varies, yielding the best overall performance on classes such as support devices, pleural effusion, and cardiomegaly. For certain clinically actionable findings such as edema and consolidation, \maira{} exhibits unsatisfactory recall, albeit with consistently high negative predictive value.
This highlights that reporting aggregate metrics alone may obscure disparate performance within findings classes or patient subgroups.

To understand whether models such as \maira{} are clinically useful, more fine-grained metrics, categories, and exemplar datasets are important, as well as evaluations in realistic use-contexts~\citep{huang2023generative}. For example, the commonly used CheXpert classes include `Pneumonia' despite it being a clinical diagnosis which should not be assessed from an image alone. Efforts such as RadCliQ are valuable towards the development of radiology-specific evaluation methods, but, as highlighted in our error analysis, the presence of incomplete or otherwise `imperfect' reports in the dataset mean correctly generated reports may still be unfairly penalised.

Reports often contain information derived from prior studies, clinical notes, accompanying laterals or other relevant imaging examinations of the patient. Models such as \maira{} and much prior work, trained to generate reports from single images, are thus forced to hallucinate this information from their limited context.
Existing commonly reported metrics do not directly quantify this failure mode. Future versions of \maira{} could include the current and previous study, thereby reducing the need to hallucinate, as demonstrated in \cite{bannur2023learning}.

\section{Conclusion}
We have presented \maira{} as a proof-of-concept for a radiology-adapted large multimodal model. Despite training on a relatively small dataset with a conventional language-modelling loss, it exhibits competitive performance with existing state-of-the-art on findings generation across a broad suite of metrics, benefiting from a domain-specific image encoder, simple training paradigm and text-based augmentation.
We believe the performance and clinical utility of \maira{} can be pushed much further by allowing the model to consider multiple images, e.g.\ priors and complementary views, and by training on larger, more diverse, and higher-quality datasets.

    \bibliographystyle{plainnat}
    \bibliography{references}  %
    \section*{Appendix}
\appendix

\section{Do we need to pre-train the adapter?} 
\label{sec:pretrain_adapter}
Prior work conducted an initial training of just the adapter layer, keeping the \ac{LLM} frozen~\citep{liu_visual_2023,li_llava-med_2023}. In \Cref{tab:ablations:stage1_effect}, we investigate this choice. We start from a baseline model with CLIP-ViT-L-336px as image encoder and Vicuna-7B as LLM. We obtain a fine-tuned adapter by training on the findings generation task. We then compare the impact of using this adapter against a randomly initialised adapter of the same size in our typical training setup. Although prior work often uses this two-stage training process, we find that initializing from the pretrained adapter causes the final performance to be significantly worse.

\begin{table}[htb]
\centering
\caption{Compare starting from a pretrained adapter vs a random one. The pretrained adapter is obtained from a separate training of the image model, adapter and LLM on the findings generation task, freezing both the LLM and the image encoder. Performance numbers are median and 95\% confidence intervals from 500 bootstrap replicates from the MIMIC-CXR test set.}
\footnotesize
\setlength{\tabcolsep}{3pt}
\begin{tabular}{@{} l l| c c @{}}\toprule
    Category & Metric & Random adapter init & Pretrained adapter init \\
    \midrule
    Lexical & 
    ROUGE-L & 28.2 {\scriptsize \textcolor{gray}{[27.8, 28.8]}}
        & 27.4 {\scriptsize \textcolor{gray}{[27.0, 27.9]}}
        \\
    & BLEU-1 & 32.8 {\scriptsize \textcolor{gray}{[32.0, 33.5]}}
        & 31.6 {\scriptsize \textcolor{gray}{[30.9, 32.3]}}
        \\
    & BLEU-4 & 12.7 {\scriptsize \textcolor{gray}{[12.2, 13.2]}} 
        & 12.1 {\scriptsize \textcolor{gray}{[11.7, 12.7]}}
        \\
    & METEOR & 30.3 {\scriptsize \textcolor{gray}{[29.8, 30.9]}}
        & 29.2 {\scriptsize \textcolor{gray}{[28.6, 29.7]}}
        \\
    \midrule
    Clinical & RadGraph-F1 & 20.2 {\scriptsize \textcolor{gray}{[19.7, 20.8]}}
        & 18.4 {\scriptsize \textcolor{gray}{[17.8, 18.9]}}
        \\
    & RG\textsubscript{ER} & 25.0 {\scriptsize \textcolor{gray}{[24.4, 25.6]}}
        & 23.0 {\scriptsize \textcolor{gray}{[22.5, 23.7]}}
        \\
    & CheXbert vector & 39.6 {\scriptsize \textcolor{gray}{[38.8, 40.4]}}
        & 35.5 {\scriptsize \textcolor{gray}{[34.6, 36.5]}}
        \\
    & RadCliQ ($\downarrow$) & 3.29 {\scriptsize \textcolor{gray}{[3.25, 3.32]}}
        & 3.42 {\scriptsize \textcolor{gray}{[3.39, 3.46]}}
        \\
    \cmidrule{2-4}
    & Macro-F1-14 & 29.4 {\scriptsize \textcolor{gray}{[27.7, 30.8]}}
        & 24.0 {\scriptsize \textcolor{gray}{[22.6, 25.4]}}
        \\
    & Micro-F1-14 & 46.4 {\scriptsize \textcolor{gray}{[45.0, 47.6]}}
        & 39.7 {\scriptsize \textcolor{gray}{[38.4, 41.0]}}
        \\
    & Macro-F1-5 & 40.7 {\scriptsize \textcolor{gray}{[38.8, 42.6]}}
        & 32.6 {\scriptsize \textcolor{gray}{[30.9, 34.3]}}
        \\
    & Micro-F1-5 & 48.1 {\scriptsize \textcolor{gray}{[46.6, 49.8]}} 
        & 40.1 {\scriptsize \textcolor{gray}{[38.2, 41.9]}} 
        \\
    \bottomrule
\end{tabular}
\label{tab:ablations:stage1_effect}
\end{table}

\section{Are gains from GPT-augmentation due to training longer?}
\label{sec:gpt_control}
We see from \Cref{tab:baseline_results_test} that adding GPT augmentation improves the performance of the model on clinical metrics. Hoewever, adding GPT paraphrased reports also increases the number of samples in the dataset and thus the training steps, as we train for the full 3 epochs even after adding the paraphrased example to our dataset. To examine whether the gains we get from using GPT-augmented data are purely due to training for a larger number of steps, we run an ablation. We create a variant dataset with the same set of samples as that from GPT-augmentation, but using the \emph{original} report rather than the GPT-paraphrased variant. In \Cref{tab:gpt_control} we compare this setting with the use of GPT-paraphrased reports.

\begin{table}[htb]
\centering
\caption{Experiment controlling for the relationship between training steps and model performance. `Control' indicates an experiment where we replace the GPT augmented reports with the original in the dataset, thus keeping the training steps constant and changing only the data. Performance numbers are median and 95\% confidence intervals from 500 bootstrap replicates from the MIMIC-CXR test set.}
\footnotesize
\setlength{\tabcolsep}{3pt}
\begin{tabular}{@{} l l| c c @{}}\toprule
    Category & Metric & \maira{} & Control \\
    \midrule
    Lexical & 
    ROUGE-L & 
        28.9 {\scriptsize\color{gray}[28.4, 29.4]} & 
        28.5 {\scriptsize\color{gray}[28.0, 28.9]}\\
    & BLEU-4 & 
        14.2 {\scriptsize\color{gray}[13.7, 14.7]} & 
        14.0 {\scriptsize\color{gray}[13.6, 14.5]}\\
    & METEOR & 
        33.3 {\scriptsize\color{gray}[32.8, 33.8]}& 
        32.7 {\scriptsize\color{gray}[32.2, 33.2]}\\
    \midrule
    Clinical & RadGraph-F1 & 
        \textbf{24.3} {\scriptsize\color{gray}[23.7, 24.8]} & 
        22.8 {\scriptsize\color{gray}[22.1, 23.3]}\\
    & RG\textsubscript{ER} & 
        \textbf{29.6} {\scriptsize\color{gray}[29.0, 30.2]} & 
        27.8 {\scriptsize\color{gray}[27.3, 28.4]}\\
    & CheXbert vector & 
        \textbf{44.0} {\scriptsize\color{gray}[43.1, 44.9]} & 
        42.8 {\scriptsize\color{gray}[42.0, 43.6]} \\
    & RadCliQ ($\downarrow$) & 
        \textbf{3.10} {\scriptsize\color{gray}[3.07, 3.14]} & 
        3.16 {\scriptsize\color{gray}[3.13, 3.19]}\\
    \cmidrule{2-4}
    & Macro-F1-14 & 
        38.6 {\scriptsize\color{gray}[37.1, 40.1]} & 
        36.7 {\scriptsize\color{gray}[35.0, 38.2]}\\
    & Micro-F1-14 & 
        \textbf{55.7} {\scriptsize\color{gray}[54.7, 56.8]} & 
        54.3 {\scriptsize\color{gray}[53.3, 55.4]}\\
    & Macro-F1-5 & 
        \textbf{47.7} {\scriptsize\color{gray}[45.6, 49.5]} & 
        45.2 {\scriptsize\color{gray}[43.3, 47.1]}\\
    & Micro-F1-5 & 
        56.0 {\scriptsize\color{gray}[54.5, 57.5]} & 
        54.4 {\scriptsize\color{gray}[53.0, 55.8]}\\
    \bottomrule
\end{tabular}
\label{tab:gpt_control}
\end{table}

\end{document}